\documentclass[reqno]{amsart}
\usepackage{mathtools}
\usepackage{color}

\def\col{black}

\def\R{{\mathbb R}}
\def\Z{{\mathbb Z}}

\def\ux{{\underline{x}}}

\def\uy{{\underline{y}}}

\def\cC{{\mathcal C}}

\def\cL{{\mathcal L}}

\def\cP{{\mathcal P}}

\def\rank{{\rm rank}}

\def\1{{\bf 1}}

\def\Z{\theta}
\def\uZ{\underline{\Z}}

\def\X0{X_0}

\def\eqnn{\begin{eqnarray*}}
\def\eeqnn{\end{eqnarray*}}
\def\eqn{\begin{eqnarray}}
\def\eeqn{\end{eqnarray}}

\def\prf{\begin{proof}}
\def\endprf{\end{proof}}

\theoremstyle{plain}
\newtheorem{theorem}{Theorem}[section]

\newtheorem{remark}[theorem]{Remark}

\numberwithin{equation}{section}

\begin{document} 

\title[Non-approximability by gradient descent in DL]
{On non-approximability of zero loss global $\cL^2$ minimizers by gradient descent in Deep Learning}

\author{Thomas Chen}
\address[T. Chen]{Department of Mathematics, University of Texas at Austin, Austin TX 78712, USA}
\email{tc@math.utexas.edu} 
\author{Patricia Mu\~{n}oz Ewald}
\address[P. Mu\~{n}oz Ewald]{Department of Mathematics, University of Texas at Austin, Austin TX 78712, USA}
\email{ewald@utexas.edu} 

\begin{abstract}
We analyze geometric aspects of the gradient descent algorithm in Deep Learning (DL), and give a detailed discussion of the circumstance that in underparametrized DL networks, zero loss minimization can generically not be attained. As a consequence, we conclude that the distribution of training inputs must necessarily be non-generic in order to produce zero loss minimizers, both for the method constructed in \cite{cheewa-2,cheewa-4}, or for gradient descent \cite{ch-7} (which assume clustering of training data). 
\end{abstract}

\maketitle

\section{Introduction and Main Results}

We analyze some basic geometric aspects of the gradient descent algorithm in Deep Learning (DL) networks. 
For some thematically related background, see for instance \cite{hanrol,grokut-1,KMTM24,lcbh,manvanzde,nonreeste,PHD20} and the references therein. 
In our previous papers \cite{cheewa-2,cheewa-4}, we gave an explicit construction of globally minimizing weights and biases for the $\cL^2$ cost in underparametrized ReLU DL networks, leading to zero loss (i.e., the value of the cost is zero). In the work at hand, we address the fact that in the underparametrized case, zero loss minimizers do generically not exist. As a consequence, we conclude that the distribution of training inputs must necessarily be non-generic to allow for zero loss minimizers, both for the method constructed in \cite{cheewa-2,cheewa-4} (which assumes clustering of training data), or for gradient descent \cite{ch-7}. 

We let the input space be given by $\R^M$, with training inputs $x_{j}^{(0)}\in \R^M$, $j=1,\dots,N$.
We assume that the outputs are given by $y_{\ell}\in\R^Q$, $\ell=1,\dots,Q$ where $Q\leq M$. We introduce the map $\omega:\{1,\dots,N\}\rightarrow\{1,\dots,Q\}$, which assigns the output label $\omega(j)$ to the $j$-th input label. That is, $x_{j}^{(0)}$ corresponds to the output $y_{\omega(j)}$. We define $\uy_\omega:=(y_{\omega(1)},\dots,y_{\omega(N)})^T\in\R^{NQ}$, where $A^T$ is the transpose of the matrix $A$. Let $N_i$ denote the number of training inputs belonging to the output vector $y_i$, $i=1,\dots,Q$. 
%We define
%\eqn 
%	\cN:=\diag\big(N_i\1_{N_i\times N_i} \,|\,i=1,\dots,Q \,\big) \;\;\in\R^{N\times N}
%\eeqn 
%where $N:=\sum_{i=1}^Q N_i$ is the total number of training inputs.

We assume that the DL network contains $L$ hidden layers, with the $\ell$-th layer defined on $\R^{M_\ell}$, and recursively determined by
\eqn
	x_j^{(\ell)} = \sigma(W_\ell x_j^{(\ell-1)} + b_\ell) \;\;\in\R^{M_\ell}
\eeqn
via the weight matrix $W_\ell\in\R^{M_\ell\times M_{\ell-1}}$, bias vector $b_\ell\in\R^{M_\ell}$, and activation function $\sigma$. We assume that $\sigma$ has a Lipschitz continuous derivative, and that the output layer 
\eqn
	x_j^{(L+1)} = W_{L+1} x_j^{(L)} + b_{L+1} \;\;\in\R^{Q}
\eeqn
contains no activation function.

We let the vector $\uZ \in\R^K$ enlist all components of all weights $W_\ell$ and biases $b_\ell$, $\ell=1,\dots,L+1$, including those in the output layer. Accordingly,
\eqn
	K = \sum_{\ell=1}^{L+1} (M_\ell M_{\ell-1}+M_\ell )
\eeqn
where we define $M_0\equiv M$ for the input layer.

In the output layer, we denote  $x_j^{(L+1)}\in \R^Q$ by $x_j[\uZ]$ for brevity, and obtain the $\cL^2$ cost as
\eqn\label{eq-cC-def-1-0}
	\cC[\ux[\uZ]] &=&
{\color{\col}
	\frac1{2N}\big|\ux[\uZ]-\uy_\omega\big|_{\R^{QN}}^2
}
	\nonumber\\
	&=& \frac1{2N}\sum_j |x_j[\uZ]-y_{\omega(j)}|_{\R^Q}^2 \,,
\eeqn 
using the notation $\ux:=(x_1,\dots,x_N)^T\in\R^{QN}$. Here, $|\bullet|_{\R^n}$ is the Euclidean norm.

\subsection{Comparison model.}
We consider the following toy model for comparison, defined by the gradient flow, 
\eqn\label{eq-uxgrad-comp-1-0}
	\partial_s \ux(s) = - \nabla_{\ux}\cC[\ux(s)]
	\;\;,\;\;
	\ux(0)=\ux^{(0)} \in\R^{QN}
\eeqn
parametrized by $s\in\R$,
or in components,
\eqn 
	\partial_s (x_j(s)-y_{\omega(j)})&=& - \frac1{N} (x_j(s)-y_{\omega(j)})
\eeqn 
for all $j=1,\dots,N$. This is trivially solvable,  
\eqn 
	x_j(s)-y_{\omega(j)} = e^{-\frac s{N}}(x_j(0)-y_{\omega(j)})
\eeqn 
with initial data $x_j(0)=x_{j}^{(0)}$. Because the right hand side converges to zero as $s\rightarrow\infty$, we find that $x_j(s)\rightarrow y_{\omega(j)}$ as $s\rightarrow\infty$ for all $j$. In particular, this yields a zero loss, global minimum of the cost, since $\cC[\ux(s)]\rightarrow0$ as $s\rightarrow\infty$.

\subsection{Gradient descent flow.}
The gradient descent algorithm seeks to minimize the cost function by use of the gradient flow for the vector of weights and biases defined by
\eqn\label{eq-uZ-graddesc-1-0}
	\partial_s \uZ(s) = -\nabla_{\uZ}\cC[\ux[\uZ(s)]] 
	\;\;,\;
	\uZ(0)=\uZ_0 \;\in\R^K \,,
\eeqn 
where the vector field $\nabla_{\uZ}\cC[\ux[\bullet]]:\R^K\rightarrow\R^K $ is Lipschitz continuous if the same holds for the derivative of the activation function $\sigma$. Accordingly, the existence and uniqueness theorem for ordinary differential equations holds for \eqref{eq-uZ-graddesc-1-0}. 
In computational applications, the initial data $\uZ_0\in\R^K$ is often chosen at random. 
Clearly, because of
\eqn\label{eq-cC-negder-1-0}
	\partial_s\cC[\ux[\uZ(s)]] 
	= - \big|\nabla_{\uZ}\cC[\ux[\uZ(s)]]\big|_{\R^K}^2
	\leq 0 
\eeqn 
the cost $\cC[\ux[\uZ(s)]]$ is monotone decreasing in $s$, and since $\cC[\ux[\uZ(s)]]\geq0$ is bounded below, the limit $\cC_*=\lim_{s\rightarrow\infty}\cC[\ux[\uZ(s)]]$ exists for any orbit $\{\uZ(s)|s\in\R\}$, and depends on the initial data, 
{\color{\col} 
$\cC_0=\cC[\ux[\uZ(0)]]$. 
}

Convergence of $\cC[\ux[\uZ(s)]]$ implies that $\lim_{s\rightarrow\infty}|\partial_s\cC[\ux[\uZ(s)]]|=0$, and therefore, $\lim_{s\rightarrow\infty}|\nabla_{\uZ}\cC[\ux[\uZ(s)]]|_{\R^K}=0$ from \eqref{eq-cC-negder-1-0}. Thus, the basic goal is to find $\cC_*=\lim_{s\rightarrow\infty}\cC[\ux[\uZ(s)]]=\cC[\ux[\uZ_*]]$ where $\uZ_*$ is a critical point of the gradient flow \eqref{eq-uZ-graddesc-1-0}, satisfying $0=-\nabla_{\uZ}\cC[\ux[\uZ_*]]$.

\begin{remark}\label{rem-nonconvZ-1-0}
	Notably,  as $s\rightarrow\infty$, neither does $\lim_{s\rightarrow\infty}\cC[\ux[\uZ(s)]]=\cC[\ux[\uZ_*]]$ imply that $\uZ(s)$ converges to $\uZ_*$, nor to any other element of $\{\uZ_{**}\in\R^K\,|\,\cC[\ux[\uZ_{**}]]=\cC[\ux[\uZ_*]]\}$, nor that $\uZ(s)$ converges at all, without further assumptions on $\cC[\ux[\bullet]]$ (for instance, of it being Morse-Bott). 
	
	Therefore, while  $\cC[\ux[\uZ(s)]]$ always converges to a stationary value of the cost function under the gradient descent flow, $\uZ(s)$ cannot generally be assumed to converge to a minimizer $\uZ_*$. 
	This is a key shortcoming of the gradient descent method, as for the training of a DL network, the main task is to find minimizing weights and biases $\uZ_*$.
\end{remark}

\begin{remark}
	As an elementary 1-dimensional example illustrating the situation addressed in Remark \ref{rem-nonconvZ-1-0}, we may consider $x[\Z]=\Z \frac{1}{\Z^2+1}$ and $\cC[x[\Z]]=\frac12 (x[\Z])^2=\frac12 \Z^2 \frac{1}{(\Z^2+1)^2}\geq0$ for $\Z\in\R$. Here, clearly, $\Z_*=0$ is a critical value and global minimizer. The gradient descent flow is determined by $\partial_s \Z(s) = - \partial_\Z \cC[x[\Z(s)]] = \Z(s) ((\Z(s))^2 -1) \frac{1}{((\Z(s))^2+1)^3}$, and one easily verifies that given any initial data with $|\Z_0|<1$, the corresponding orbit converges, $\lim_{s\rightarrow\infty}\Z(s)=\Z_*=0$. 

	On the other hand, given any initial data with $|\Z_0|>1$, the corresponding orbit diverges, $\lim_{s\rightarrow\infty}|\Z(s)|=\infty$, while nevertheless, $\lim_{s\rightarrow\infty}x[\Z(s)]=0$, and therefore, $\lim_{s\rightarrow\infty}\cC[x[\Z(s)]]=0=\cC[x[\Z_*]]$. 
	This is because $|\partial_\Z \cC[x[\Z]]|\sim\frac1{|\Z|^3}$ for $|\Z|\gg1$, and one straightforwardly verifies that for $\Z\gg1$, the solution of $\partial_s \Z(s)\sim \frac1{(\Z(s))^3}$ has the asymptotic behavior $\Z(s)\sim s^{\frac14}\rightarrow\infty$ as $s\rightarrow\infty$. The case for $\Z\ll-1$ is similar.
\end{remark}

\subsection{Dynamics of $\ux(s):=\ux[\uZ(s)]$}
Next, we note that $\cC[\ux[\uZ(s)]]$ depends on $\uZ(s)$ only through its dependence on $\ux[\uZ(s)]$. 
{\color{\col}
Thus, defining the Jacobi matrix
}
\eqn
	{\color{\col}
	D[\uZ]
	}
	&:=&\Big[\frac{\partial x_j[\uZ]}{\partial \Z_\ell}\Big]_{j=1,\dots,N\;\ell=1,\dots,K}
	\nonumber\\
	&=&
	\left[
	\begin{array}{ccc}
		\frac{\partial x_1[\uZ]}{\partial \Z_1} & \cdots & \frac{\partial x_1[\uZ]}{\partial \Z_K } \\
		\cdots & \cdots & \cdots \\
		\frac{\partial x_{N}[\uZ]}{\partial \Z_1} & \cdots & \frac{\partial x_{N}[\uZ]}{\partial \Z_K }
	\end{array}\right]
	\;\;\;
	\in \R^{QN\times K}
\eeqn
and writing $\ux(s):=\ux[\uZ(s)]$ for brevity, we find that the gradient descent flow for $\uZ(s)$ induces the following flow for $\ux(s)\in\R^{QN}$, 
\eqn\label{eq-ux-ODE-1-0}
	\partial_s\ux(s) &=&D[\uZ(s)] \, \partial_{s}\uZ
	\nonumber\\
	&=&
	- D[\uZ(s)] \nabla_{\uZ}\cC[\ux[\uZ(s)]]
	\nonumber\\
	&=&
	- D[\uZ(s)] D^T[\uZ(s)] \nabla_{\ux}\cC[\ux[\uZ(s)]]
\eeqn 
Passing to the second line, we used \eqref{eq-uZ-graddesc-1-0}.
Here, the matrix $D[\uZ(s)]D^T[\uZ(s)]\in\R^{QN\times QN}$ is positive semi-definite;
{\color{\col}
it corresponds to the neural tangent kernel \cite{JGH18}.
}
In the special case when it is strictly positive definite and thus invertible, \eqref{eq-ux-ODE-1-0} is the gradient flow for $\ux(s)$ in the metric on $\R^{QN}$ defined by the metric tensor $(D[\uZ(s)]D^T[\uZ(s)])^{-1}$.
Our main results in this paper address the similarity or dissimilarity in the qualitative behavior between solutions to \eqref{eq-ux-ODE-1-0} and the comparison model \eqref{eq-uxgrad-comp-1-0}, depending on this invertibility condition.

\subsubsection{The overparametrized case}
In the overparametrized situation where $K\geq QN$, we have the following result.

\begin{theorem}
Assume that $\ux[\uZ_*]$ is a stationary solution,
\eqn\label{eq-statsol-1-0}
	0 = - D[\uZ_*] D^T[\uZ_*] \nabla_{\ux}\cC[\ux[\uZ_*]]
\eeqn
Then, it corresponds to a global minimum of the $\cL^2$ cost,
\eqn
	\cC[\ux[\uZ_*]] = 0 \,,
\eeqn
if and only if $\nabla_{\ux}\cC[\ux[\uZ_*]]=0$.

A necessary condition for $\nabla_{\ux}\cC[\ux[\uZ_*]]=0$ to follow from \eqref{eq-statsol-1-0} is that
\eqn\label{eq-DDT-fullrank-1-0}
	\rank(D[\uZ_*]D^T[\uZ_*])=QN
\eeqn 
has full rank. This in turn is only possible if $K\geq QN$ which means that the DL network is overparametrized. 

Moreover, if there exist $s_0\geq0$ and $\lambda>0$ such that $D[\uZ(s)] D^T[\uZ(s)]>\lambda$ for all $s\geq s_0$ (so that in particular, $\rank(D[\uZ(s)] D^T[\uZ(s)])=QN$) along the orbit $\uZ(s)$, the solution of \eqref{eq-ux-ODE-1-0} converges to the global minimizer for any initial condition $\ux(0)\in\R^{QN}$.
\end{theorem}

\prf
In components,  $\nabla_{\ux}\cC[\ux[\uZ_*]]=0$ is explicitly given by
\eqn 
	\frac1{N}(x_j[\uZ_*]-y_{\omega(j)}) = 0 \;\;\;\forall j\in\{1,\dots,N\} \,.
\eeqn 	
Therefore, $\nabla_{\ux}\cC[\ux[\uZ_*]]=0$ is equivalent to $x_j[\uZ_*]=y_{\omega(j)}$ for all $j$, and thus holds if and only if $\cC[\ux[\uZ_*]]=0$.

We recall that $D[\uZ_*]\in\R^{QN\times K}$ where $K$ is the total number of parameters contained in all weights and biases. Therefore, $\rank(D[\uZ_*] D^T[\uZ_*])\leq \min\{QN,K\}$, and for $D[\uZ_*] D^T[\uZ_*] $ to have full rank $QN$, it is necessary that $QN\leq K$. But this means that the DL network is overparametrized.

Finally, if there exists $s_0\geq 0$ such that $D[\uZ(s)] D^T[\uZ(s)]>\lambda$ for a positive constant $\lambda>0$ and all $s\geq s_0$, then  
\eqn
	\partial_s \cC[\ux(s)]&=&-(\nabla_{\ux}\cC[\ux(s)])^TD[\uZ(s)] D^T[\uZ(s)] \nabla_{\ux}\cC[\ux(s)]
	\nonumber\\
	&\leq&- \lambda |\nabla_{\ux}\cC[\ux(s)]|_{\R^{QN}}^2
	\nonumber\\
	&=&- 2\frac \lambda {N} \cC[\ux(s)]
\eeqn
for all $s>s_0$. %, where $N_*:=\min\{N_j\,|\,j=1,\dots,Q\}$.
Therefore, 
{\color{\col}
$\lim_{s\rightarrow\infty}\cC[\ux(s)]\leq \lim_{s\rightarrow\infty}e^{-2\frac\lambda {N}(s-s_0)}\cC[\ux(s_0)]=0$.
} Because $\cC[\ux(s)]$ is a convex function of $\ux(s)-\uy_{\omega}$,  this implies that for any arbitrary initial data $\ux(0)=\ux_0\in\R^{QN}$, the solution of \eqref{eq-ux-ODE-1-0} converges to the global minimizer $\ux_*=\lim_{s\rightarrow\infty}\ux(s)$ which  satisfies $\ux_{*}-\uy_{\omega}=0$.
\endprf

\begin{remark}
	We note that while $\ux_*=\lim_{s\rightarrow\infty}\ux(s)=\lim_{s\rightarrow}\ux[\uZ(s)]$ converges in the above situation, the vector of weights and biases $\uZ(s)$ itself nevertheless does not necessarily converge.
\end{remark}

\subsubsection{The underparametrized case}
In the underparametrized situation where $K<QN$, we have the following result.

\begin{theorem}\label{prp-underparam-1-0}
	Assume that $K<QN$, and that $\uZ(s)$, $s\in\R$, is an orbit of the gradient descent flow \eqref{eq-uZ-graddesc-1-0}. Denote by $\cP[\uZ(s)]$ the projector, orthogonal with respect to the Euclidean inner product on $\R^{QN}$, onto the range of $D[\uZ(s)] D^T[\uZ(s)]$ where $\rank(D[\uZ(s)] D^T[\uZ(s)])\leq K$ (the latter is not assumed to be constant in $s$), and let $\cP^\perp[\uZ(s)]:=\1_{QN\times QN}-\cP[\uZ(s)]$ denote its complement.
Then, 
\eqn 
	\partial_s\ux(s) &=&- \cP[\uZ(s)] (D[\uZ(s)] D^T[\uZ(s)] \nabla_{\ux}\cC[\ux[\uZ(s)]])
	\nonumber\\
	\cP^\perp[\uZ(s)]\partial_s\ux(s) &=& 0 
\eeqn 
has the structure of a constrained dynamical system.
In particular,
\eqn\label{eq-def-cP-1-0}
	\cP[\uZ(s)]=D[\uZ(s)] (D^T[\uZ(s)] D[\uZ(s)])^{-1} D^T[\uZ(s)] \,,
\eeqn 
if $\rank(D[\uZ(s)] D^T[\uZ(s)])=K$ is maximal.
Let $\uZ_*$ be an arbitrary stationary point of the cost function, with $\nabla_{\uZ}\cC[\ux[\Z_*]]=0$, and $\rank(D[\uZ_*] D^T[\uZ_*])\leq K$. 
%Let $\cP[\uZ_*]$ denote the orthoprojector onto the range of $D[\uZ_*] D^T[\uZ_*]$, and $\cP^\perp[\uZ_*]=\1_{QN\times QN}-\cP[\uZ_*]$. 
Then,
\eqn\label{eq-cPnabC-1-0}
	0 = \cP[\uZ_*]\nabla_{\ux}\cC[\ux[\uZ_*]] \,.
\eeqn 
In particular, the local extremum of the cost function at $\uZ_*$ is attained at
\eqn\label{eq-cCmin-GD-1-0}
	\cC[\ux[\uZ_*]] = \frac N2 
	\big|\cP^\perp[\uZ_*]\nabla_{\ux}\cC[\ux[\uZ_*]] \big|_{\R^{QN}}^2 
\eeqn 
where $\rank(\cP^\perp[\uZ_*])\geq QN-K$.
\end{theorem}

\prf
Due to being a symmetric matrix, $D[\uZ(s)] D^T[\uZ(s)]=R^T\Lambda R$ for any given $s\in\R$ (where we notationally suppress the dependence of $R$ and $\Lambda$ on $\uZ(s)$ for  brevity), where $\Lambda\geq0$ is diagonal and $R\in SO(QN)$. Then, letting $P_\Lambda$ denote the projector obtained from replacing all nonzero entries of $\Lambda$ by $1$, we have $\cP[\uZ(s)]=R^T P_\Lambda R$. From $P_\Lambda \Lambda=\Lambda=\Lambda P_\Lambda$ follows that
\eqn\label{eq-DDT-commut-1-0}
	D[\uZ(s)] D^T[\uZ(s)] &=& \cP[\uZ(s)] D[\uZ(s)] D^T[\uZ(s)] 
	\nonumber\\
	&=& D[\uZ(s)] D^T[\uZ(s)] \cP[\uZ(s)] \,.
\eeqn 
In other words, $[D[\uZ(s)] D^T[\uZ(s)],\cP[\uZ(s)] ]=0$ commute, and from
\eqn\label{eq-DDT-PDDTP-1-0}
	D[\uZ(s)] D^T[\uZ(s)] =\cP[\uZ(s)]D[\uZ(s)] D^T[\uZ(s)] \cP[\uZ(s)] \,,
\eeqn 
the ranges and kernels of $D[\uZ(s)] D^T[\uZ(s)]$ and $\cP[\uZ(s)] $ coincide, for every $s\in\R$.

If $\rank(D[\uZ(s)] D^T[\uZ(s)])=K$ is maximal, then the matrix $D^T[\uZ(s)] D[\uZ(s)]\in\R^{K\times K}$ is invertible, as a consequence of which the expression \eqref{eq-def-cP-1-0} for the orthoprojector $\cP[\uZ(s)] $ is well-defined. 

It follows from \eqref{eq-ux-ODE-1-0} and \eqref{eq-DDT-commut-1-0} that
\eqn\label{eq-Pux-ODE-1-0}
	\partial_s\ux(s)  
	&=&
	- \cP[\uZ(s)] (D[\uZ(s)] D^T[\uZ(s)] \nabla_{\ux}\cC[\ux[\uZ(s)]]) \,,
\eeqn 
and as a consequence,
\eqn 
	\cP^\perp[\uZ(s)]\partial_s\ux(s) = 0 \,.
\eeqn 
It follows from \eqref{eq-ux-ODE-1-0} that $\partial_s \uZ(s)=0$ implies $\partial_s\ux(s)=0$. 

{\color{\col}
Let $\uZ_*$ denote a stationary point for \eqref{eq-uZ-graddesc-1-0}, with $\rank(D[\uZ_*] D^T[\uZ_*])\leq K$, so that clearly, $\rank(\cP^\perp[\uZ_*])\geq QN-K$.
}
Then,
\eqn
	0 &=&  - D[\uZ_*] D^T[\uZ_*] \nabla_{\ux}\cC[\ux[\uZ_*]]
%	\nonumber\\
%	&=& - D[\uZ_*] D^T[\uZ_*] \cP[\uZ_*] \nabla_{\ux}\cC[\ux[\uZ_*]] 
\eeqn
from which follows that
\eqn 
	\cP[\uZ_*]\nabla_{\ux}\cC[\ux[\uZ_*]] = 0 \,,
\eeqn 
due to \eqref{eq-DDT-PDDTP-1-0}.
Then, 
\eqn
	\cC[\uZ_*] &=& \frac N2 |\nabla_{\ux}\cC[\ux[\uZ_*]] |^2
	\nonumber\\
	&=&
	\frac N2 \Big(\big|\cP[\uZ_*]\nabla_{\ux}\cC[\ux[\uZ_*]] \big|_{\R^{QN}}^2
	+ \big|\cP^\perp[\uZ_*]\nabla_{\ux}\cC[\ux[\uZ_*]] \big|_{\R^{QN}}^2 \Big)
	\nonumber\\
	&=&
	\frac N2 \big|\cP^\perp[\uZ_*]\nabla_{\ux}\cC[\ux[\uZ_*]] \big|_{\R^{QN}}^2
\eeqn  
as claimed. 
\endprf

\subsubsection{Comparison with constructive minimizers from \cite{cheewa-2}}

The map $\ux:\R^K\rightarrow\R^{QN}$, $\uZ\mapsto\ux[\uZ]$ from the space of parameters to the output space is determined by the training inputs $\{\,(x_{j,i}^{(0)})_{i=1}^{N_j}\, \}_{j=1}^Q$. Accordingly, the map $\ux:\R^K\rightarrow\R^{QN}$ is generic if the distribution of training inputs $\{\,(x_{j,i}^{(0)})_{i=1}^{N_j}\, \}_{j=1}^Q$ is generic (i.e., it is highly random and unstructured).

We therefore arrive at the following main result.
%, for which we assume that $N_j=\frac{N}{Q}$, $j=1,\dots,Q$, so that the cost function \eqref{eq-cC-def-1-0} matches the one used in \cite{cheewa-2}. The cost function in \cite{cheewa-2} accommodates for varying values of $N_j$; adapting our results to this more general case is straightforward, but we refrain from it here for simplicity and brevity of exposition.

\begin{theorem}
\label{cor-generic-underpar-1-0-0}
Zero loss minimizers of underparametrized ReLU DL networks 
%as considered in \cite{cheewa-2,cheewa-4,ch-7} 
do not exist for generic distributions of training data. 

%Non-genericness of the distribution of training data in  underparametrized ReLU DL networks (for instance, via clustering properties, as in \cite{cheewa-2,cheewa-4,ch-7}) is a necessary condition for the existence of global zero loss minimizers $\uZ_*$, as constructed explicitly in \cite{cheewa-2,cheewa-4}, or via gradient descent in \cite{ch-7}.
\end{theorem}

\prf
The ReLU activation function $\sigma$ acts  component-wise by the ramp function $(\xi)_+=\max\{0,\xi\}$ for $\xi\in\R $. Suitably smoothing the latter in an $\epsilon$-neighborhood of the origin for an arbitrary small $\epsilon>0$, we obtain $\sigma_\epsilon$, which we assume to be monotone increasing, and to have a Lipschitz continous derivative. Accordingly, the corresponding gradient vector field $\nabla_\ux\cC[\ux[\uZ]]$ and the matrix $D[\uZ]$ are Lipschitz continuous in $\uZ$. Therefore,
{\color{\col} 
Theorem \ref{prp-underparam-1-0}
} 
can be applied to the flow generated by it. 
{\color{\col}
	For generic training inputs, the right hand side of \eqref{eq-cCmin-GD-1-0} is strictly positive, due to the nonzero rank of $\cP^\perp[\uZ_*]$; accordingly, zero loss does not occur. 
}
We conclude that zero loss minimizing weights and biases for the $\cL^2$ cost in underparametrized ReLU DL networks do not exist, and can not be approximated via the gradient descent flow, if the distribution of training inputs is generic.
\endprf 

The DL network studied in \cite{cheewa-2,cheewa-4,ch-7} is underparametrized, with $M=M_\ell=Q=L$ $\forall \ell$ and $K=(Q+1)^3+(Q+1)^2<QN$ (respectively, $\ll QN$), and uses the ReLU activation function.
The minimizers obtained in \cite{cheewa-2,cheewa-4,ch-7} are robust under a small deformation of $\sigma$ to a monotone increasing $\sigma_\epsilon$ (in particular, they involve no derivatives of the activation function). Accordingly, the construction given in  \cite{cheewa-2,cheewa-4} with  $\sigma$ replaced by $\sigma_\epsilon$ yields a degenerate zero loss minimum of the cost function. 
%On the other hand, Theorem \ref{prp-underparam-1-0} implies that the gradient descent algorithm generically does not produce a global minimum in the underparametrized situation. 
In \cite{cheewa-2,cheewa-4,ch-7}, the training data is clustered, and hence non-generic.
Therefore, the existence of zero loss minimizers, constructed explicitly for underparametrized ReLU DL in \cite{cheewa-2,cheewa-4}, and via gradient flow in \cite{ch-7}, is not in contradiction with the above.

$\;$\\
\noindent
{\bf Acknowledgments:} 
{\color{\col}
We thank the anonymous referee for very useful comments.
}
T.C. thanks Cy Mayor for helpful discussions.
T.C. gratefully acknowledges support by the NSF through the grant DMS-2009800, and the RTG Grant DMS-1840314 - {\em Analysis of PDE}. P.M.E. was supported by NSF grant DMS-2009800 through T.C. 
\\


\begin{thebibliography}{99}

		 

\bibitem{ch-7}
T. Chen,
{\em Derivation of effective gradient flow equations and dynamical truncation of training data in Deep Learning}.
https://arxiv.org/abs/2501.07400

\bibitem{cheewa-1}
T. Chen, P. Mu\~{n}oz Ewald, 
{\em Geometric structure of shallow neural networks and constructive $\cL^2$ cost minimization}.
https://arxiv.org/abs/2309.10370


\bibitem{cheewa-2}
T. Chen, P. Mu\~{n}oz Ewald, 
{\em Geometric structure of Deep Learning networks and construction of global $\cL^2$ minimizers}.
https://arxiv.org/abs/2309.10639

\bibitem{cheewa-4}
T. Chen, P. Mu\~{n}oz Ewald,
{\em Interpretable global minima of deep ReLU neural networks on sequentially separable data.}
arxiv.org/abs/2405.07098 

\bibitem{hanrol}
B. Hanin, D. Rolnick, 
{\em How to start training: The effect of initialization and architecture}, Conference on Neural Information Processing Systems (NeurIPS) 2018.

\bibitem{grokut-1} 
P. Grohs, G. Kutyniok, (eds.)
{\em Mathematical aspects of deep learning},
Cambridge University Press, Cambridge (2023).

{\color{\col}
\bibitem{JGH18} A. Jacot, F. Gabriel, C. Hongler. {\em Neural tangent kernel: Convergence and generalization in neural networks}. Advances in neural information processing systems, {\bf 31} (2018).
}


\bibitem{KMTM24} K. Karhadkar, M. Murray, H. Tseran, G. Montufar. {\em  Mildly overparameterized relu networks have a favorable loss landscape} (2024). arXiv:2305.19510.

\bibitem{lcbh}
Y. LeCun, Y. Bengio, and G. Hinton. 
{\em Deep Learning}. Nature, 521:436–521 (2015).

\bibitem{manvanzde}
S.S.  Mannelli, E. Vanden-Eijnden, L. Zdeborov\'{a},
{\em Optimization and Generalization of Shallow Neural Networks with Quadratic Activation Functions},
NIPS'20: Proceedings of the 34th International Conference on Neural Information Processing Systems, December 2020. Article No.: 1128, pp. 13445-13455.

\bibitem{nonreeste}
M. Nonnenmacher, D. Reeb, I. Steinwart,
{\em Which Minimizer Does My Neural Network Converge To ?}
Machine Learning and Knowledge Discovery in Databases. Research Track: European Conference, ECML PKDD 2021, Bilbao, Spain, September 13–17, 2021, Proceedings, Part III Sep 2021, pp 87–102.
 
 
\bibitem{PHD20} 
V. Papyan, X.Y. Han, D.L. Donoho, 
{\em Prevalence of neural collapse during the terminal phase of deep learning training}. Proceedings of the National Academy of Sciences, {\bf 117} (40), 24652-24663 (2020).

	
\end{thebibliography}
\end{document}